\documentclass[10pt,twocolumn,letterpaper]{article}

\usepackage{cvpr}
\usepackage{times}
\usepackage{epsfig}
\usepackage{graphicx}
\usepackage{amsmath}
\usepackage{amssymb}
\usepackage{bm}
\usepackage{array}

\usepackage{color}
\usepackage{booktabs}

\usepackage[pagebackref=true,breaklinks=true,letterpaper=true,colorlinks,bookmarks=false]{hyperref}

\cvprfinalcopy 

\def\cvprPaperID{4760} 

\ifcvprfinal\fi
\begin{document}

\title{Fully-Featured Attribute Transfer}

\author{De~Xie$^1$, Muli Yang$^1$, Cheng~Deng$^{1,}$\thanks{Corresponding author}, Wei~Liu$^2$, Dacheng~Tao$^3$\\
	$^1$School of Electronic Engineering, Xidian University, Xi'an 710071, China\\
	$^2$Tencent AI Lab, Shenzhen, China\\
	$^3$UBTECH Sydney AI Centre, SIT, FEIT, University of Sydney, Australia\\
	\{xiede.xd, muliyang.xd, chdeng.xd\}@gmail.com, wliu@ee.columbia.edu,  dacheng.tao@sydney.edu.au
}
\maketitle

%

\begin{abstract}
	Image attribute transfer aims to change an input image to a target one with expected attributes, which has received significant attention in recent years. However, most of the existing methods lack the ability to de-correlate the target attributes and irrelevant information, i.e., the other attributes and background information, thus often suffering from blurs and artifacts. To address these issues, we propose a novel Attribute Manifold Encoding GAN (AME-GAN) for fully-featured attribute transfer, which can modify and adjust every detail in the images. Specifically, our method divides the input image into image attribute part and image background part on manifolds, which are controlled by attribute latent variables and background latent variables respectively. Through enforcing attribute latent variables to Gaussian distributions and background latent variables to uniform distributions respectively, the attribute transfer procedure becomes controllable and image generation is more photo-realistic. Furthermore, we adopt a conditional multi-scale discriminator to render accurate and high-quality target attribute images. Experimental results on three popular datasets demonstrate the superiority of our proposed method in both performances of the attribute transfer and image generation quality.

\end{abstract}
\vspace{-0.3cm}
\section{Introduction}
\begin{figure}[!t]
	\centering
	\includegraphics[width=8.3cm,height=8.35cm]{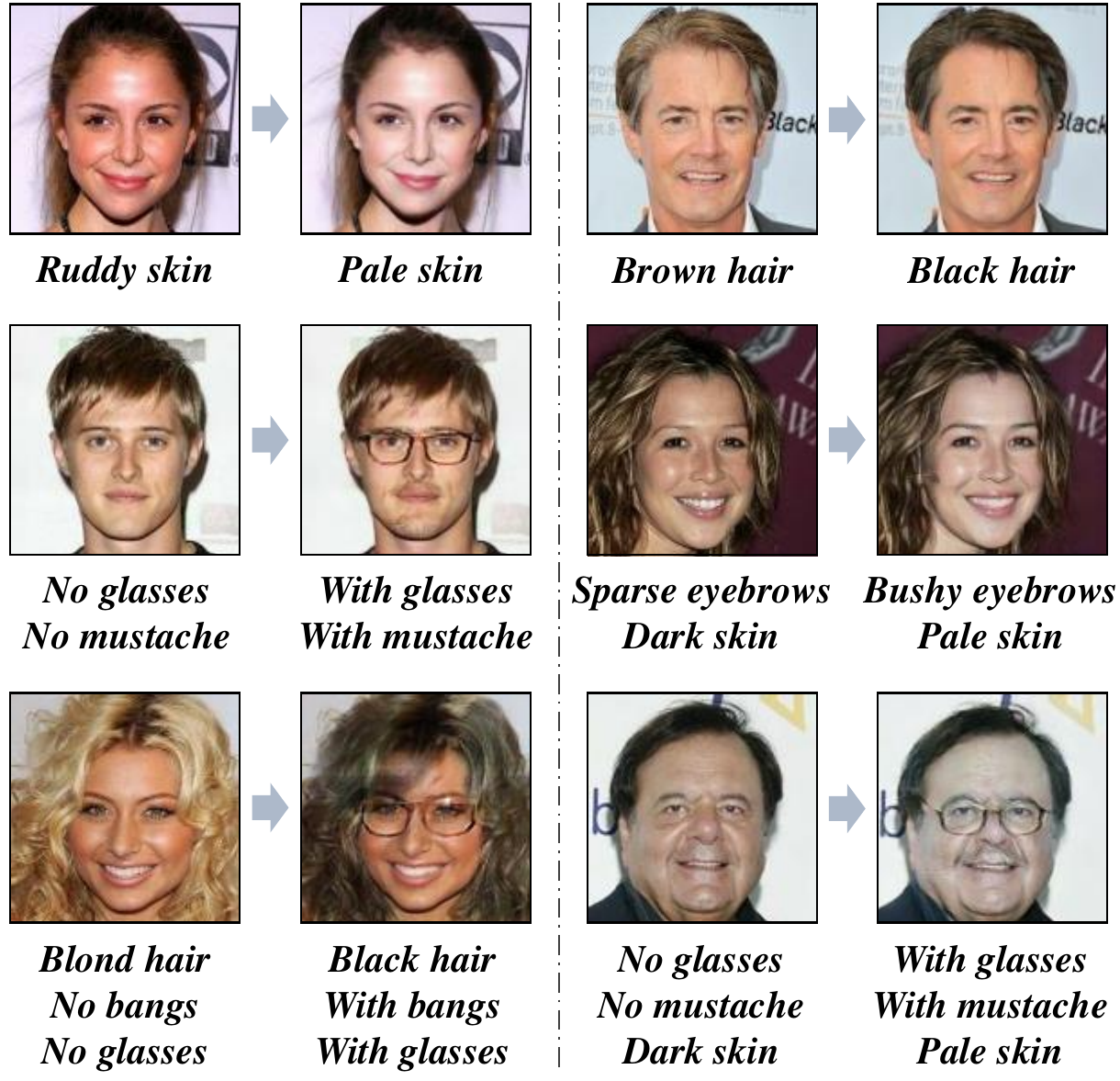}
	\vspace{0.0cm}
	\caption{The illustration of our attribute transfer results for facial images, which contain single-attribute transfer and multiple-attribute transfer.}
	\vspace{-0.5cm}
	\label{fig:1}
\end{figure}

The task of image attribute transfer aims at manipulating a source image to possess single or multiple target attributes (\eg, eyeglasses, bangs, and mustaches of facial images. See Fig.~\ref{fig:1}). 
The desired results should not only contain the attributes of interest, but also keep other attributes and background invariant, such as the identity of a facial image.
Early attempts~\cite{gatys2015texture,gatys2015neural,zhang2013eyeglasses,zhu2014recover} often address this problem in a regression fashion by explicitly designing the loss functions, such as pixel-wise losses,
and the target attribute is thereby learned by inputting a pair of images that only differ from the desired attribute (\eg, facial images of the same person with or without eyeglasses).
Although such methods can precisely capture the target attributes from paired images, the paired images are actually difficult to attain and are thus unaffordable in real-world applications.

With the success of Generative Adversarial Networks (GANs)~\cite{goodfellow2014generative} in diverse computer vision tasks~\cite{chen2018sketchygan,Li_2018_CVPR,ledig2017photo,zhang2017age,xu2017attngan,chen2018cartoongan,zhao2018modular}, GAN based methods~\cite{CycleGAN2017,xiao2017GeneGAN,wang2017tag,xiao2018dna,StarGAN2018,Xiao_2018_ECCV} are springing up and have exhibited promising performance on image attribute transfer.
By implicitly defining the loss function with a discriminator network which is trained in an adversarial manner with the generator network, GAN based methods are more flexible in the unpaired training scenario. 
CycleGAN~\cite{CycleGAN2017} harnesses a cycle structure to perform attribute transfer as a cross-domain image-to-image translation task, which is able to transfer a single attribute from a source image to the target one.
GeneGAN~\cite{xiao2017GeneGAN} learns an attribute subspace by encoding an input image into an attribute-relevant part and an attribute-irrelevant part, which can transfer a single attribute through toggling the two encoding parts in the learned subspace.
However, these methods are only able to transfer a single attribute at a time, which are inflexible for the multiple-attribute scenario.

For multiple-attribute transfer, some recent studies~\cite{lample2017fader,wang2017tag,xiao2018dna,Xiao_2018_ECCV} argue that multiple attributes of an image can be disentangled into different parts in a latent encoding space.
However, these methods have a limited capability to separate the attributes and background in the latent encoding space, which renders mutual interference during attribute transfer and often bring blurs and artifacts in the generated images.
StarGAN~\cite{StarGAN2018} designs a multi-domain cycle structure and carries out multiple-attribute transfer by inputting an image and its corresponding binary attribute label. However, it simply regards attribute transfer as a multi-domain image-to-image translation task, which lacks the ability to model attributes effectively and cannot disentangle multiple attributes, thereby failing to control the procedure of attribute transfer.

In this paper, we propose an \emph{Attribute Manifold Encoding GAN} (AME-GAN) to realize fully-featured attribute transfer. 
One can regard an image with attributes naturally lies on a high-dimensional manifold $\mathcal{M}_I$, which can be further divided into two submanifolds for the attributes and background separation based on the nonlinear separability of manifold. 
On the basis of this observation, we elaborately devise an attribute decoder and a background decoder to project attribute latent variables and background latent variables to the two submanifolds, \ie, $\mathcal{M}_A$ and $\mathcal{M}_B$, which can effectively separate attributes and background information.
Specifically, i.i.d.~Gaussian distributions and uniform distributions are respectively imposed on attribute latent variables and background ones to describe the independence of different attributes and the consistency of background information.
In this way, we can modify and adjust every detail in the images with realistic generated results through recombining the target attributes information and the background information on manifold.
To ensure the accuracy and high-resolution of transfered images, we design a conditional multi-scale discriminator to classify attributes and distinguish images from different scales.
 
Our contributions can be summarized in four-fold:\vspace{-0.2cm}
\begin{itemize}
	\item As we know, we address the multiple-attribute transfer task via manifold learning for the first time, and separate image attributes and background information based on the nonlinear separability of manifold.\vspace{-0.2cm}
	\item We devise a novel framework for fully-featured attribute transfer. Especially, we design a conditional multi-scale discriminator to generate accurate and high-resolution transferred images.\vspace{-0.2cm}
	\item We employ Gaussian and uniform distributions to describe the independence of different attributes and the consistency of background information, respectively, which can not only preserve background information but also transfer attributes flexibly and realistically.\vspace{-0.2cm}
	\item Comprehensive experimental results demonstrate the superiority of our method in both the accuracy of attribute transfer and the quality of image generation under various scenarios.\vspace{-0.2cm}
\end{itemize}

\section{Related Work}
\subsection{Image Attribute Transfer}
Recently, numerous image attribute transfer methods have been proposed, which can be mainly divided into two types: optimization-based methods and learning-based methods.
The main idea of optimization-based methods, such as CNAI~\cite{li2016convolutional} and DFI~\cite{upchurch2017deep}, is to find the difference between an input image and the target one with desired attributes, and then design a loss function to model the difference between them.
In contrast, learning-based methods have received increasing attention in recent years due to their flexibility and expandability.
Variational Autoencoder (VAE) and Generative Adversarial Network (GAN) based methods~\cite{goodfellow2014generative,kingma2013auto,liu2017unsupervised} are designed to learn an attribute latent representation and a corresponding decoder, which can realize attribute transfer through modifying the latent representation of input images to be close to that of the target images, followed by the decoder.
IcGAN~\cite{perarnau2016invertible} combines cGAN~\cite{mirza2014conditional} with an encoder.
It can sample from a uniform distribution to obtain the latent representation irrelevant to input image attributes, and then encode the target attributes into this latent representation and decode it to obtain the image with target attributes.
Some image translation methods, such as CycleGAN~\cite{CycleGAN2017}, are able to transfer images across image domains.

Other attempts explore exchanging latent codes in the latent attribute space of the input images to implement attribute transfer.
GeneGAN~\cite{xiao2017GeneGAN} swaps a specific attribute between two given images by recombining the information of their latent representations.
TD-GAN~\cite{wang2017tag} and DNA-GAN~\cite{xiao2018dna} also exchange attribute latent code blocks between a given pair of images to generate hybrid images.
ELEGANT~\cite{Xiao_2018_ECCV} applies a U-Net structure~\cite{ronneberger2015u_net} on the basis of DNA-GAN for high-resolution image generation. 
StarGAN~\cite{StarGAN2018} and AttGAN~\cite{he2017arbitrary} realize attribute transfer by introducing an attribute classification loss.
Moreover, StarGAN~\cite{StarGAN2018} designs a conditional attribute transfer network to learn attributes in a cyclic process, while AttGAN~\cite{he2017arbitrary} devises an encoder-decoder architecture to model the relationship between the latent representations and the attributes.
Both StarGAN and AttGAN are designed for learning multiple attributes simultaneously.
However, they fail to frame attribute information sufficiently by only  concatenating attribute labels to image latent variables, which is limited to decoupling the correlation of different attributes and leads to unrealistic results. 

Our proposed method can cope with both image attributes and background information by mapping images to a latent space and encode them to the corresponding latent representations.
The independence of different attributes and the consistency of background information are respectively preserved through imposing Gaussian and uniform distributions on the latent representations.
These latent representations are then projected to the manifolds conditional on the target attributes through two deconvolutional generators to handle the attributes and the background information separately and flexibly.
This scheme can make the procedure of attribute transfer more explicable and provide an effective solution to all the problems mentioned above.
\subsection{Generative Adversarial Network}
Generative Adversarial Network (GAN) has been widely used for image generation tasks~\cite{bodla2018semi,lu2018attribute,ma2018gan,zhang2018generative}.
GAN comprises two components: a generator $G$ and a discriminator $D$.
The generator $G$ captures the distribution of training samples and learns to generate new samples imitating the training ones, and the discriminator $D$ tries to distinguish the generated samples from the training ones.
$G$ and $D$ are trained adversarially with each other using a \emph{min-max game} strategy.
The objective function of GAN is given as follows:
\begin{small}
	\begin{equation}
	{\min\limits_{G}}{\max\limits_{D}} {\mathbb{E}_{x{\sim}p_{data}\!({x})}} \! \left[\mathrm{log}D(x)\right] \!+\!  {\mathbb{E}_{x{\sim}p_z\!{({z})}}}\!\!\left[\mathrm{log}\big(1-D(G(x))\big)\right]\!,
	\label{1}
	\end{equation}
\end{small}where $z$ denotes a vector randomly sampled from a prior distribution $p_z{(z)}$, and $p_{data}(x)$ is the data distribution.
There are many variations of GAN.
DCGAN~\cite{radford2015unsupervised} adopts deconvolutional and convolutional neural networks to implement $G$ and $D$, respectively, which promotes the application of GANs in many image generation tasks.
cGAN~\cite{mirza2014conditional} modifies GAN from unsupervised learning to semi-supervised learning by feeding the conditional variables into the data and achieves promising performance.

In this paper, we design a conditional multi-scale discriminator. Through introducing multi-scale conditional discriminative information to the discriminator, we can significantly improve the permormance of GANs and therefore generate accurate images with high quality.
\section{Attribute Manifold Encoding GAN}
\subsection{Overview}
\label{section:overview}
In the attribute transfer task, an image with attributes can be divided into two components: the attribute part and the background part. The background part should remain unchanged while flexibly changing the attribute part.
Suppose that an image $\bm{I}$ with multiple attributes lies on a high-dimensional manifold $\mathcal{M}_I$, on which traversing along a certain direction could achieve attribute transfer while preserving the image background information.
Mathematically, the mapping process can be formulated as follows:
\begin{equation}
	\bm{I} = f_I(\bm{r}_a, \bm{r}_b),
	\label{2}
\end{equation} 
where $\bm{r}_a$ and $\bm{r}_b$ are the representations of image attributes and background, respectively. $f_I$ is the mapping from $\bm{r}_a$ and $\bm{r}_b$ to $\mathcal{M}_I$. In order to customize attributes, we assume $\bm{r}_a$ lies on a submanifold of $\mathcal{M}_I$, denoted by $\mathcal{M}_A$, which can be controlled with the attribute label $\bm{x}$ and the attribute latent variable $\bm{l}_a$. It can be formulated as:
\begin{equation}
	\bm{r}_a = f_A(\bm{l}_a, \bm{x}),
	\label{3}
\end{equation} 
where $f_A$ is the mapping from $\bm{l}_a$ and $\bm{x}$ to $\mathcal{M}_A$.
For the fully-featured attribute transfer, the different attributes should be independently identically distributed.
According to Variational Autoencoder (VAE)~\cite{kingma2013auto}, we impose i.i.d.~Gaussian distributions on the attribute latent variable $\bm{l}_a$ and apply different means $\bm{m}_{x}$ and variances $\bm{v}_{x}$ generated from the target attribute label $\bm{x}$ to represent different attributes.
Thus, Eq.~\eqref{3} can be rewritten as follows:
\begin{equation}
	\bm{r}_a = f_A({\bm{l}_a}\cdot{\bm{v}_{x}} + \bm{m}_{x}).
	\label{4}
\end{equation} 
By doing so, the independence and heterogeneity of different attributes can be described by i.i.d.~Gaussian distributions and their means and variances, which is efficient and reasonable for fully-featured attribute transfer.

\begin{figure}[!t]
	\centering
	\includegraphics[width=8.6cm]{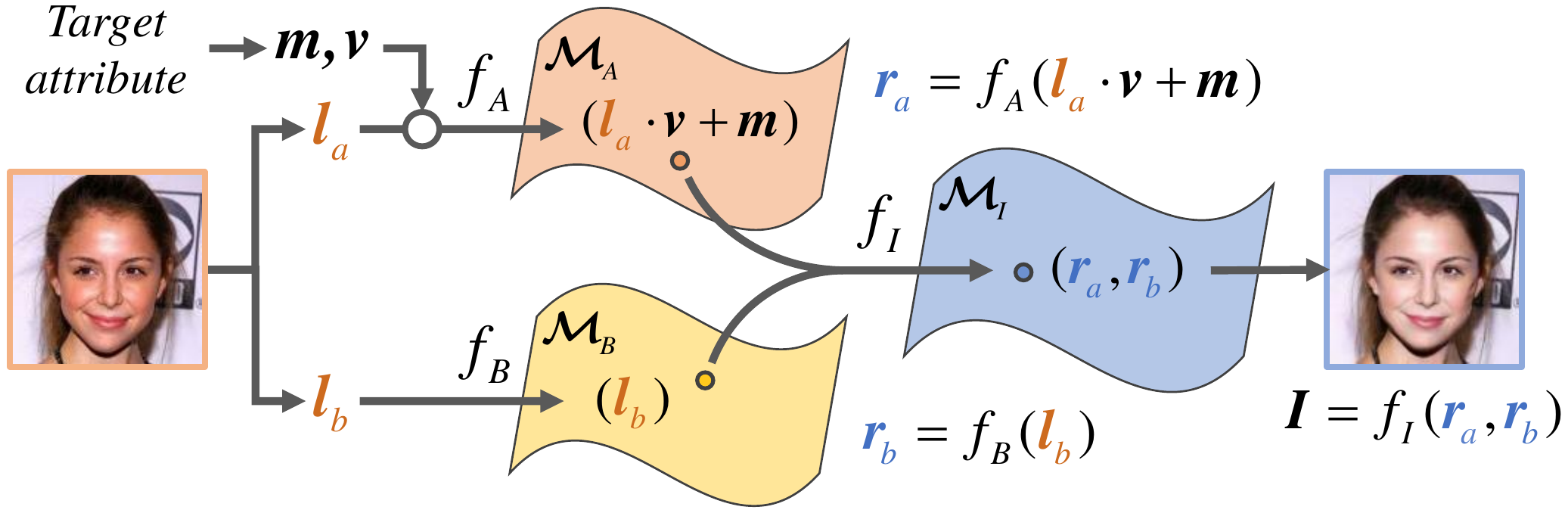}
	\vspace{0cm}
	\caption{The pipeline of the proposed approach. 
	}
	\vspace{-0.4cm}
	\label{fig:2}
\end{figure}
Regarding the background part representation, $\bm{r}_b$ can be assumed to lie on another submanifold of $\mathcal{M}_I$, denoted by $\mathcal{M}_B$, and is controlled by the background latent variable $\bm{l}_b$, which can be denoted as:
\begin{equation}
	\bm{r}_b = f_B(\bm{l}_b),
	\label{5}
\end{equation} 
where $f_B$ is the mapping from $\bm{l}_b$ to $\mathcal{M}_B$.
In order to keep the consistency of background information, inspired by CAAE~\cite{zhang2017age}, we take advantage of the uniformity of a uniform distribution and impose it on the background latent variable $\bm{l}_b$, which can render the background information smooth on $\mathcal{M}_B$ and thus generate undistorted images.
Fig.~\ref{fig:2} illustrates the pipeline of the proposed approach.
\begin{figure*}[!t]
	\centering
	\includegraphics[width=15.0cm]{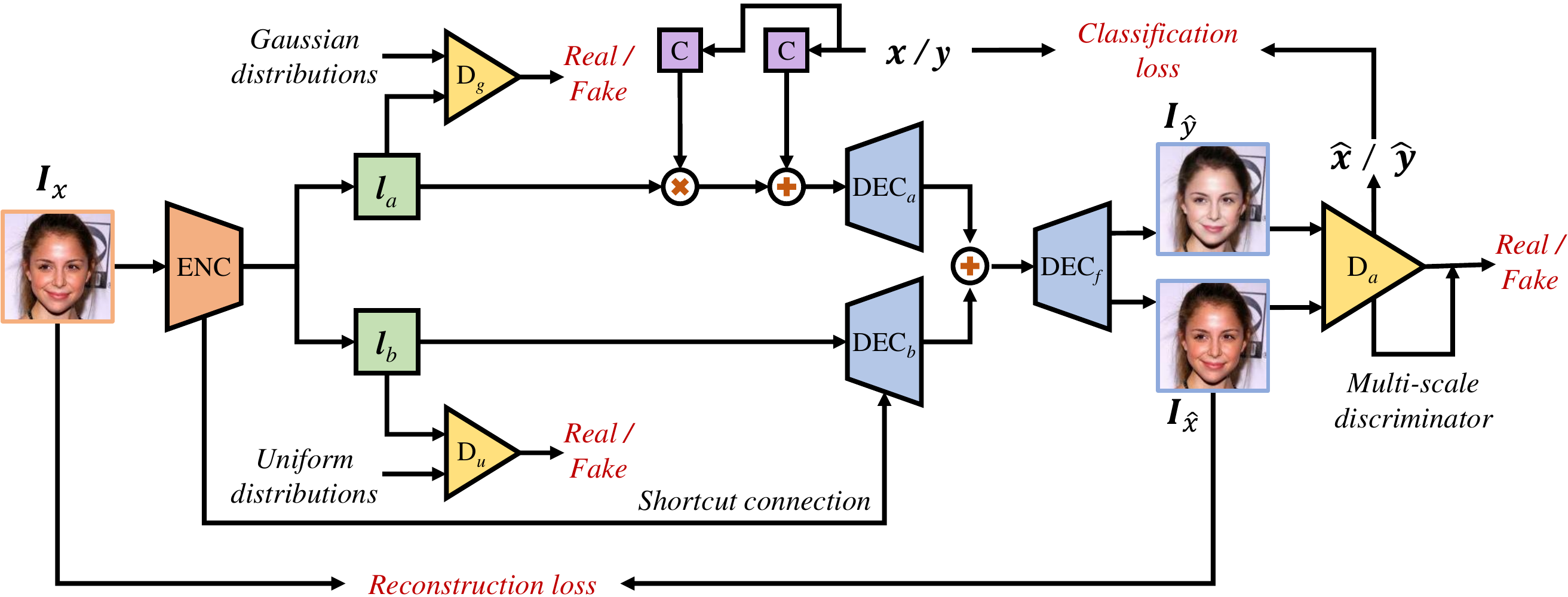}
	\vspace{0.1cm}
	\caption{The flowchart of the proposed framework, which comprises three main components, \ie,  Encoder, Decoder, and Discriminator. Given an image as input, Encoder generates two latent variables $\bm{l}_a$ and $\bm{l}_b$ for the attribute part and the background part respectively. Decoder modifies $\bm{l}_a$ by the target attribute label, which is then coupled with $\bm{l}_b$ for synthesizing an image with target attributes on manifolds.
	Discriminator imposes Gaussian and uniform distributions on $\bm{l}_a$ and $\bm{l}_b$ respectively, and moreover, encourages the generated image to be photo-realistic and reliable towards target attributes.}
	\vspace{-0.45cm}
	\label{fig:3}
\end{figure*}
\subsection{Framework}
We propose a novel framework for fully-featured attribute transfer using manifold leaning. The whole architecture is illustrated in Fig.~\ref{fig:3}. It contains three blocks: 
1) an image encoder block $\mathrm{ENC}$ that encodes an image to the attribute latent variable $\bm{l}_a$ and the background latent variable $\bm{l}_b$;
2) a decoder block, including three parts: an attribute decoder $\mathrm{DEC}_a$, a background decoder $\mathrm{DEC}_b$, and a fusion decoder $\mathrm{DEC}_f$, which map $(\bm{l}_a, \bm{x})$ to $\mathcal{M}_A$, $\bm{l}_b$ to $\mathcal{M}_B$, and $(\bm{r}_a, \bm{r}_b)$ to $\mathcal{M}_I$, respectively;
3) a discriminator block, including three parts: a Gaussian discriminator $\mathrm{D}_g$, a uniform discriminator $\mathrm{D}_u$ and an attribute discriminator $\mathrm{D}_a$.
$\mathrm{D}_g$ and $\mathrm{D}_u$ regularize $\bm{l}_a$ to Gaussian distributions, and $\bm{l}_b$ to uniform distributions, respectively.
$\mathrm{D}_a$ is a conditional multi-scale discriminator, which ensures attribute transferred images more accurate and of higher resolution through attribute classification and multi-scale image information discrimination. 

\noindent\textbf{Encoder.} Given an image $\bm{I}_{x}$ with $n$ binary attributes ${\bm{x}} = (x_1, x_2, \dots, x_n)$ and target attributes ${\bm{y}} = (y_1, y_2, \dots, y_n)$, we can obtain the attribute latent variable $\bm{l}_a$ and the background latent variable $\bm{l}_b$, which can be written as:
\begin{equation}
	(\bm{l}_a, \bm{l}_b) = \mathrm{ENC}(\bm{I}_{x}).
	\label{6}
\end{equation}

\noindent\textbf{Decoder.} In order to map $({\bm{y}}, \bm{l}_a)$ to $\mathcal{M}_A$ and $\bm{l}_b$ to $\mathcal{M}_B$, we design the image attribute decoder $\mathrm{DEC}_a$ and the image background decoder $\mathrm{DEC}_b$, which are expected to learn $f_A$ and $f_B$, respectively.
For $\mathrm{DEC}_a$, firstly, two extra groups of convolutional layers are designed to learn the mean $\bm{m}_{y}$ and variance $\bm{v}_{y}$ of target attributes. The attribute representation $\bm{r}_{a}^{y}$, which lies on the submanifold $\mathcal{M}_A$, can be formulated as:
\begin{equation}
	\bm{r}_{a}^{y} = \mathrm{DEC}_a\big(\theta \cdot (\bm{l}_a \cdot \bm{v}_{y} + \bm{m}_{y})\big),
	\label{7}
\end{equation}
where $\theta$ is a hyper-parameter (set to 1.0 during training) that can control the intensity of attribute transfer.
For $\mathrm{DEC}_b$, the background representation $\bm{r}_b$, which lies on the submanifold $\mathcal{M}_B$, can be depicted as:
\begin{equation}
	\bm{r}_b = \mathrm{DEC}_b(\bm{l}_b).
	\label{8}
\end{equation}
After obtaining $\bm{r}_a^y$ and $\bm{r}_b$, $\mathrm{DEC}_f$ is designed to combine them and map them to $\mathcal{M}_I$ to generate an attribute transferred image, which can be denoted as:
\begin{equation}
	\bm{I}_y = \mathrm{DEC}_f(\bm{r}_{a}^{y} + \bm{r}_b).
	\label{9}
\end{equation}
We introduce a reconstruction loss to measure the correctness of attribute learning, which ensures that the information of attributes is preserved in $\bm{r}_{a}^{y}$.
The reconstruction loss can be written as:
\begin{equation}
	\mathcal{L}_{recon} = \mathbb{E}_{\bm{y}=\bm{x}}\big[{\parallel}\bm{I}_y - \bm{I}_x{\parallel}_1\big].
	\label{10}
\end{equation}

\noindent\textbf{Discriminator.} The Gaussian discriminator $\mathrm{D}_g$ and the uniform discriminator $\mathrm{D}_u$ aim to force $\bm{l}_a$ to approach Gaussian distributions and $\bm{l}_b$ uniform distributions, respectively.
Suppose that $\bm{g}$ obeys a set of Gaussian distributions ($\bm{g} \sim N(0, 1)$) and $\bm{z}$ obeys a set of uniform distributions ($\bm{z} \sim U(-1, 1)$). The objective functions of $\mathrm{D}_g$ and $\mathrm{D}_u$ can be formulated as:
\begin{equation}
	\begin{array}{ll}
		\mathcal{L}_{adv}^g= &\mathbb{E}_{\bm{g} \sim N}[\mathrm{log}D(\bm{g})] + \\
		&\mathbb{E}_{\bm{l}_a \sim p(\bm{l}_a)}[\mathrm{log}(1-D(\bm{l}_a))],
		\label{11}
	\end{array}
\end{equation}
\begin{equation}
	\begin{array}{ll}
		\mathcal{L}_{adv}^u = &\mathbb{E}_{\bm{z} \sim U}[\mathrm{log}D(\bm{z})] + \\
		&\mathbb{E}_{\bm{l}_b \sim p(\bm{l}_b)}[\mathrm{log}(1-D(\bm{l}_b))].
		\label{12}
	\end{array}
\end{equation}

\begin{figure*}[!t]
	\centering
	\includegraphics[width=17.0cm,height=10.3cm]{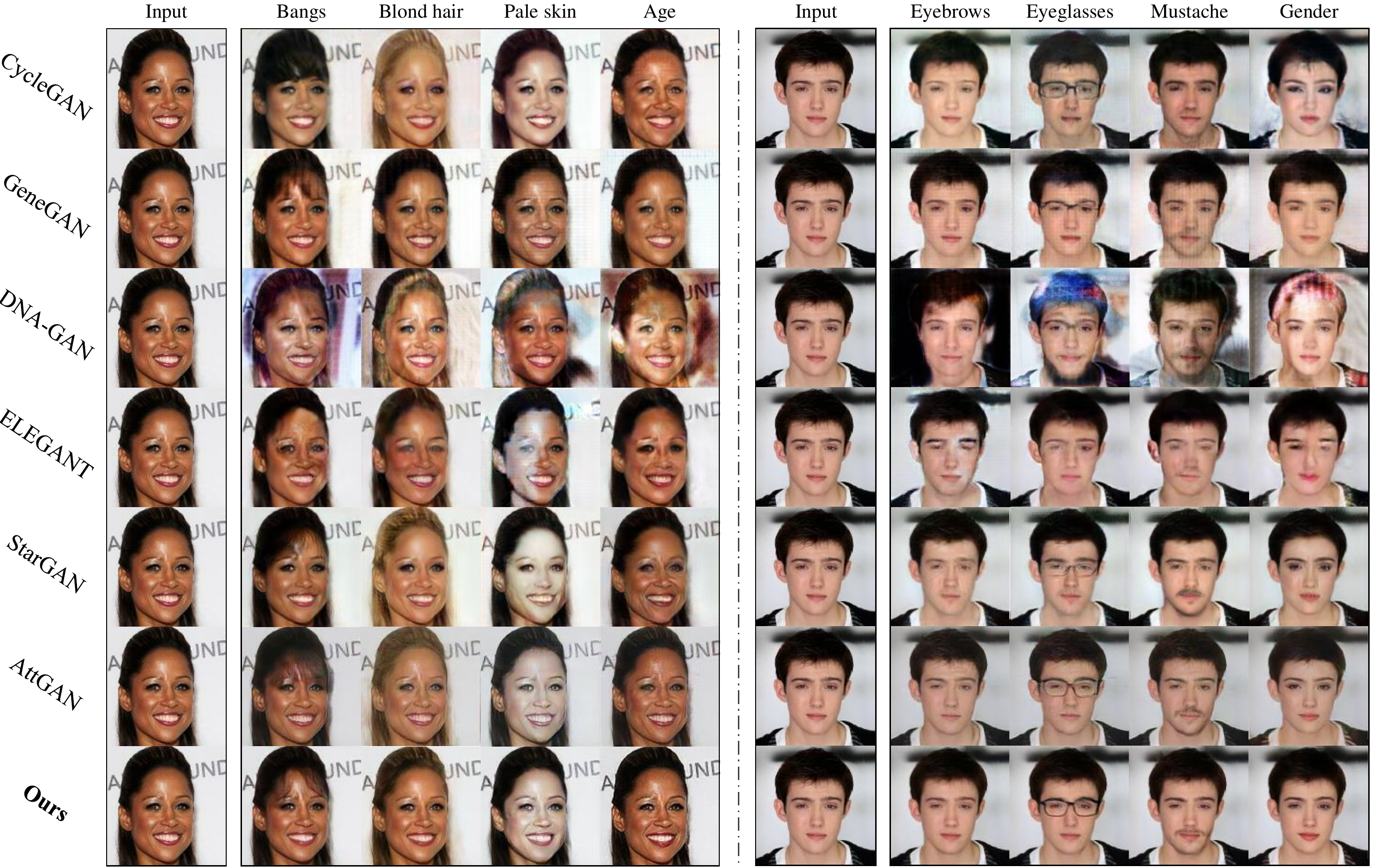}
	\vspace{0cm}
	\caption{Single-attribute transfer results on the CelebA dataset.}
	\vspace{-0.4cm}
	\label{fig:4}
\end{figure*}

For the attribute discriminator $\mathrm{D}_a$, we adopt a conditional multi-scale discriminator architecture to deal with the two tasks, \ie, generated images discrimination and attribute classification.
For the first task, suppose that $\{\bm{d}_1^I, \bm{d}_2^I, \dots, \bm{d}_m^I\}$ are the output of $m$ different layers of $\mathrm{D}_a$ with $\bm{I}$ as input.
The multi-scale objective function of $\mathrm{D}_a$ can be framed as:
\begin{equation}
	\begin{array}{ll}
		\mathcal{L}_{adv}^a = &\mathbb{E}_{\bm{I}_x \sim p(\bm{I}_x)}[\mathrm{log}\sum\limits_{i=1}^{m}{\gamma_i{\bm{d}_i^{I_x}}}] + \\
		&\mathbb{E}_{\bm{I}_y \sim p(\bm{I}_y)}[\mathrm{log}(1-\sum\limits_{i=1}^{m}{\gamma_i{\bm{d}_i^{I_y}}})],
		\label{13}
	\end{array}
\end{equation}
where $\gamma_i$ is the weight of multi-scale information $\bm{d}_i^{I_{\ast}}$.
For the attribute classification task, we adopt binary cross entropy loss to ensure efficient attribute learning.
Suppose that $\bm{c} = (c_1, c_2, \dots, c_n)$ is the predicted attribute label.
The attribute classification objective function of $\mathrm{D}_a$ can be written as:
\begin{equation}
	\mathcal{L}_{cls}^a = \sum_{i=1}^{n}{-y_i{\mathrm{log}}c_i-(1-y_i){\mathrm{log}(1-c_i)}}.
	\label{14}
\end{equation}
Combining $\mathcal{L}_{adv}^a$ with $\mathcal{L}_{cls}^a$, we can obtain the overall objective function of the attribute discriminator $\mathrm{D}_a$.


\subsection{Objective Function}
The final objective function of our method can be obtained by combining Eq.~\eqref{10}, Eq.~\eqref{11}, Eq.~\eqref{12}, Eq.~\eqref{13}, and Eq.~\eqref{14}:
\begin{equation}
	\mathcal{L}_{all} = \mathcal{L}_{recon}+\mathcal{L}_{adv}^g+\mathcal{L}_{adv}^u+\mathcal{L}_{adv}^a+\mathcal{L}_{cls}^a.
\end{equation}
The encoder, decoder and discriminator blocks are trained alternately by minimizing $\mathcal{L}_{all}$ until convergence.

\subsection{Extension to Controllable Attribute Transfer}
After the training procedure, we can customize attribute transfer with the trained encoder and decoder.
Given an image $\bm{I}_u$ with attributes $\bm{u}$ and the target attributes $\bm{v}$, firstly, the image attribute latent variable $\bm{l}_a$ and the background latent variable $\bm{l}_b$ can be obtained according to Eq.~\eqref{6}, and eventually the transferred image $\bm{I}_v$ with the target attributes $\bm{v}$ can be achieved by:
\begin{equation}
	\bm{I}_v = {\mathrm{DEC}_f}\left(\mathrm{DEC}_a\big(\theta \cdot (\bm{l}_a \cdot \bm{v}_{v} + \bm{m}_{v})\big) + \mathrm{DEC}_b(\bm{l}_b)\right),
	\label{16}
\end{equation}
where $\bm{v}_{v}$ and $\bm{m}_{v}$ are the variances and means of the target attributes generated by the binary label of $\bm{v}$.
In addition, the intensity of attribute transfer can be controlled by the value of hyper-parameter $\theta$.
\begin{table*}[h]
	\caption{Single-attribute transfer accuracy on the CelebA dataset.}
	\vspace{0.05cm}
	\begin{center}
		\scalebox{0.95}{
		\begin{tabular}{cccccccccc}
			\toprule
			Method & Bangs & Blond hair & Pale skin & Age & Eyebrows & Eyeglasses & Mustache & Gender & Average\\
			\midrule
			CycleGAN & 0.483 & 0.502 & 0.344 & 0.695 & 0.266 & 0.553 & 0.109 & 0.538 & 0.436 \\
			GeneGAN & 0.279 & 0.122 & 0.101 & 0.814 & 0.093 & 0.479 & 0.084 & 0.471 & 0.305\\
			DNA-GAN & 0.139 & 0.223 & 0.203 & 0.798 & 0.114 & 0.471 & 0.062 & 0.333 & 0.293\\
			ELEGANT & 0.397 & 0.313 & 0.232 & 0.858 & 0.166 & 0.337 & 0.092 & 0.479 & 0.359\\
			StarGAN & \bf{0.864} & \bf{0.686} & 0.426 & 0.908 & 0.263 & 0.938 & 0.156 & 0.853 & 0.637\\
			AttGAN & 0.513 & 0.638 & 0.651 & 0.895 & 0.367 & 0.962 & 0.278 & 0.967 & 0.659\\
			\bf{Ours} & 0.825 & 0.585 & \bf{0.713} & \bf{0.944} & \bf{0.458} & \bf{0.978} & \textbf{0.328} & \textbf{0.975} & \textbf{0.726}\\
			\bottomrule
		\end{tabular}}
	\end{center}
	\vspace{-0.3cm}
	\label{tab:cls}
\end{table*}
\begin{table*}[h]
	\caption{Perceptual similarity errors of single-attribute transfer results using LPIPS metric (v0.1) on the CelebA dataset.}
	\vspace{0.05cm}
	\begin{center}
		\scalebox{0.95}{
		\begin{tabular}{cccccccccc}
			
			\toprule
			Method & Bangs & Blond hair & Pale skin & Age & Eyebrows & Eyeglasses & Mustache & Gender & Average \\
			\midrule
			CycleGAN & 0.167 & 0.185 & 0.138 & 0.100 & 0.101 & 0.167 & 0.118 & 0.164 & 0.142 \\
			GeneGAN & 0.101 & 0.076 & \textbf{0.060} & 0.069 & 0.069 & 0.091 & 0.093 & 0.097 & 0.082\\
			DNA-GAN & 0.220 & 0.173 & 0.198 & 0.212 & 0.225 & 0.264 & 0.275 & 0.241 & 0.226\\
			ELEGANT & 0.097 & 0.130 & 0.215 & 0.106 & 0.143 & 0.103 & 0.129 & 0.109 & 0.129\\
			StarGAN & 0.131 & 0.157 & 0.170 & 0.104 & 0.108 & 0.130 & 0.113 & 0.114 & 0.128\\
			AttGAN & 0.110 & 0.129 & 0.123 & 0.077 & 0.079 & 0.110 & 0.082 & 0.096 & 0.101\\
			\bf{Ours} & \textbf{0.067} & \textbf{0.072} & 0.080 & \bf{0.054} & \bf{0.057} & \bf{0.065} & \textbf{0.055} & \textbf{0.062} & \textbf{0.064}\\
			\bottomrule
		\end{tabular}}
	\end{center}
	\vspace{-0.6cm}
	\label{tab:iqa}
\end{table*}

\section{Experiments}
\subsection{Datasets and Implementations}
\noindent\textbf{Datasets.} We evaluate our method on three datasets: CelebA~\cite{liu2015deep}, Caltech-UCSD-Birds 200-2011 (CUB)~\cite{wah2011caltech}, and Paintings dataset~\cite{CycleGAN2017}.
The CelebA dataset is composed of 202,599 facial images with 40 attributes, and 5 landmark locations. We use the aligned and cropped version and choose eight distinguishable attributes, \ie, Bangs, Blond hair, Pale skin, Age, Eyebrows, Eyeglasses, Mustache, and Gender for all comparison methods.
The CUB dataset contains 11,788 images from 200 different types of birds and 312 annotated attributes. We combine all the color attributes and select four strong visual impact attributes: Blue, Yellow, Black and Red.
The Paintings dataset is collected from Flickr and WikiArt, which contains five styles, \ie, Photo, Monet, Van Gogh, Cezanne and Ukiyo-e.

\noindent\textbf{Implementation details.} Our method is compared with the following six latest methods: CycleGAN, GeneGAN, DNA-GAN, ELEGANT, StarGAN and AttGAN, which achieve state-of-the-art performance for attribute transfer.
Our propose model is trained on an NVIDIA TITAN X GPU and implemented with the TensorFlow toolkit. The encoder $\mathrm{ENC}$ is equipped with five layers of Conv-BatchNorm-LeakyRelu blocks, and Deconv-InstanceNorm-Relu blocks are employed in the three decoders.
Especially, we adopt the U-Net structure for $\mathrm{DEC}_b$ that skips connection with $\mathrm{ENC}$.
$\mathrm{D}_u$ and $\mathrm{D}_g$ are both equipped with two layers of $1\times1$ Conv-BatchNorm-LeakRelu blocks followed by a fully-connected layer. 
Attribute discriminator $\mathrm{D}_a$ uses five layers of Conv-BatchNorm-LeakRelu blocks followed by two fully-connected layers for the discriminative branch and the classification branch, respectively.
All networks are trained with Adam~\cite{kingma2014adam} initialized with the learning rate set to 0.0002, ${\beta}_1$ 0.5 and ${\beta}_2$ 0.999.
All the input images are normalized into the range $[-1, 1]$ and all the comparison methods are under the same experimental setting.
\begin{figure}[!]
	\centering
	\includegraphics[height=6.2cm]{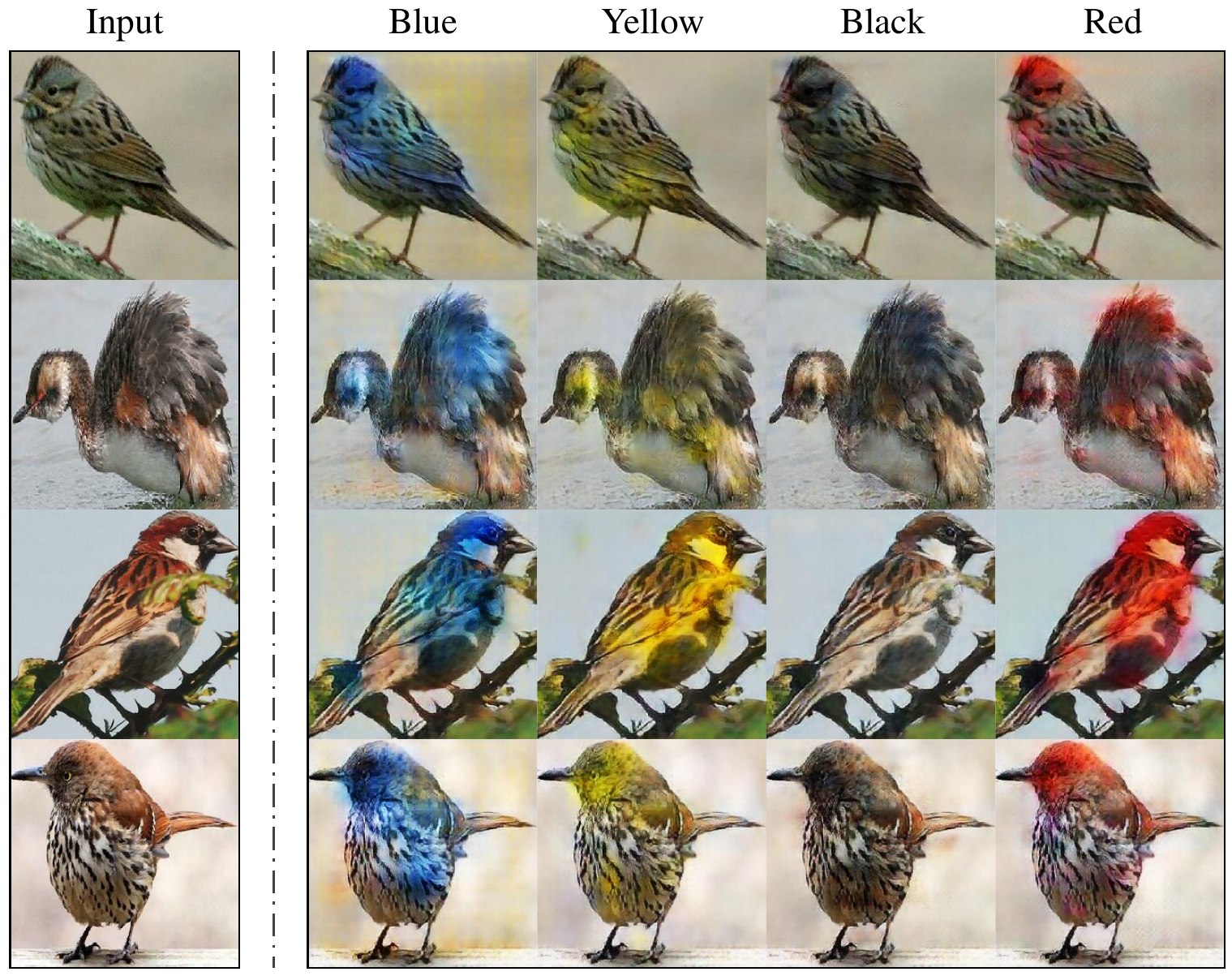}
	\vspace{0.1cm}
	\caption{Single-attribute transfer results of our method on the CUB dataset.}
	\vspace{-0.5cm}
	\label{fig:5}
\end{figure}
\begin{figure*}[!t]
	\centering
	\includegraphics[width=17.0cm,height=10.5cm]{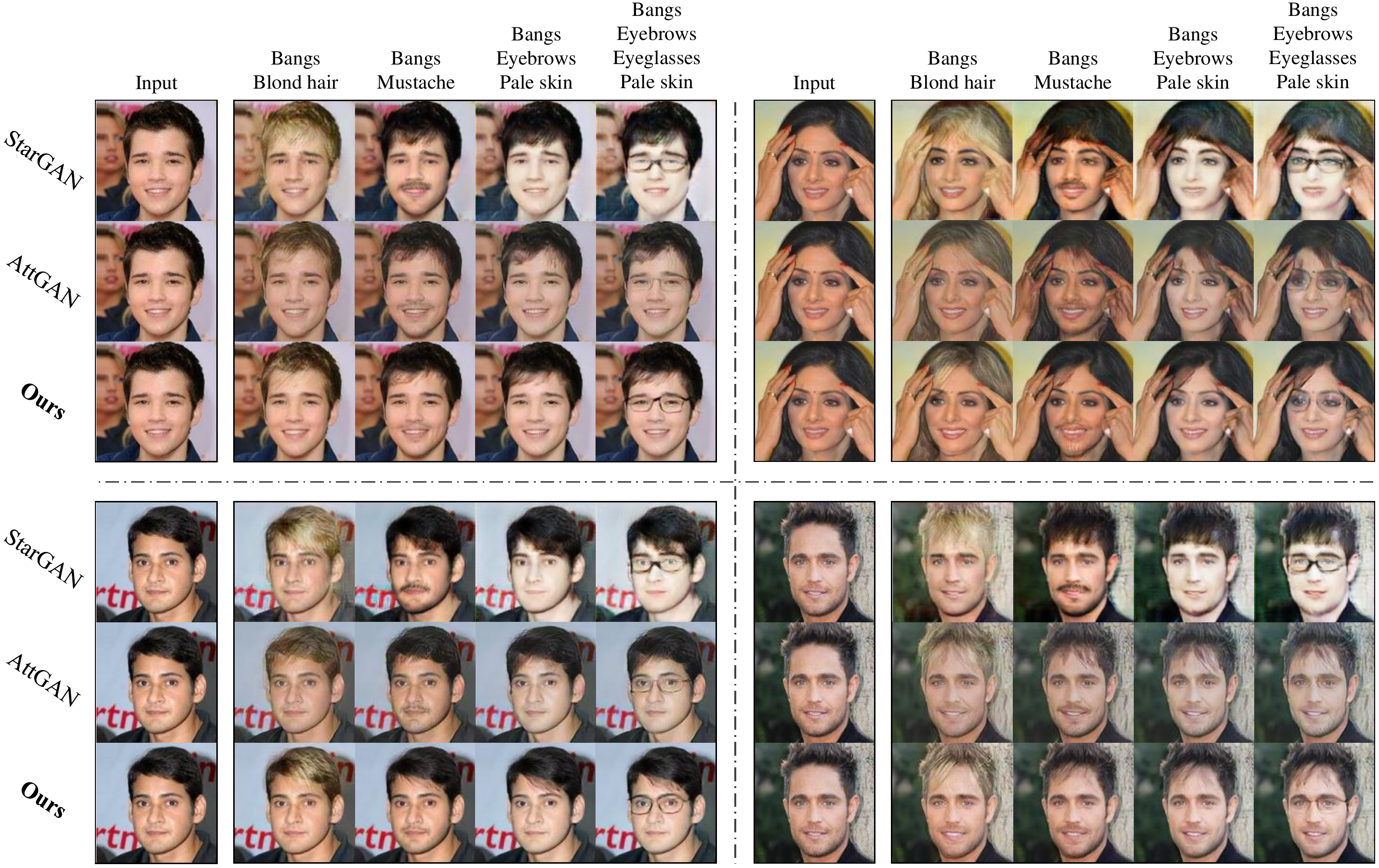}
	\vspace{0cm}
	\caption{Multiple-attribute transfer results on the CelebA dataset.}
	\vspace{-0.1cm}
	\label{fig:6}
\end{figure*}
\begin{figure*}[!t]
	\centering
	\includegraphics[width=17.0cm,height=6.0cm]{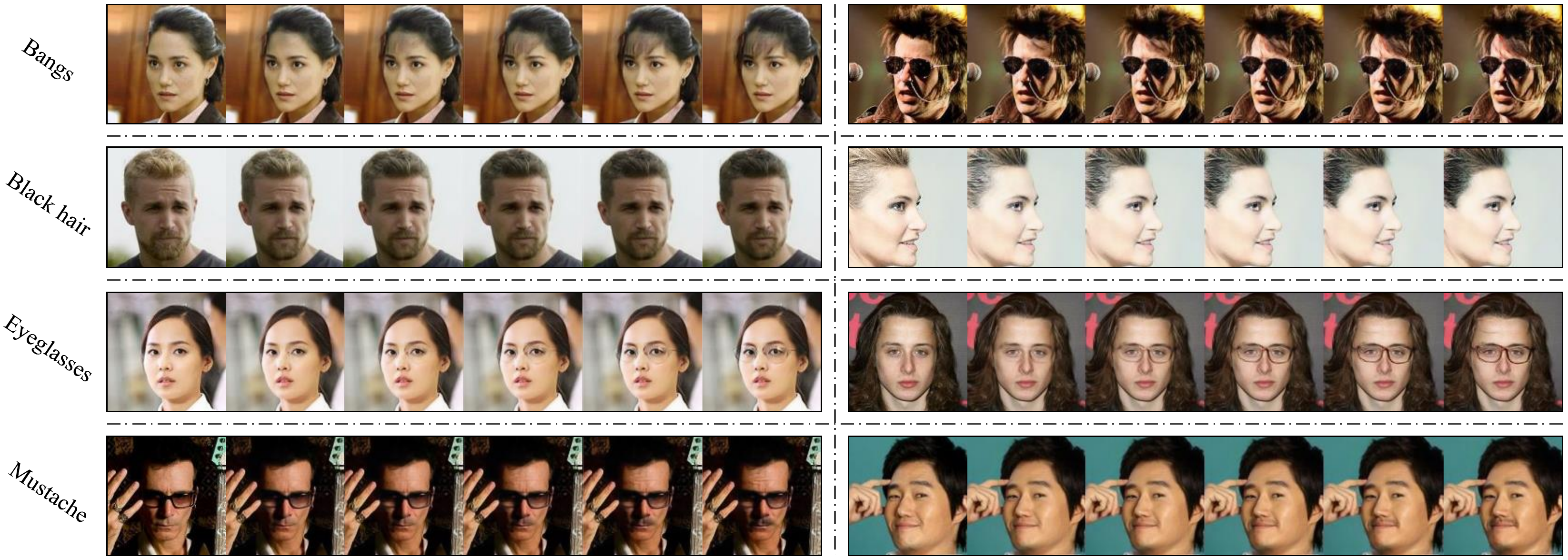}
	\vspace{-0cm}
	\caption{Controllable attribute transfer results of our method on the CelebA dataset.}
	\vspace{-0.4cm}
	\label{fig:7}
\end{figure*}
\begin{figure*}[!t]
	\centering
	\includegraphics[width=15cm]{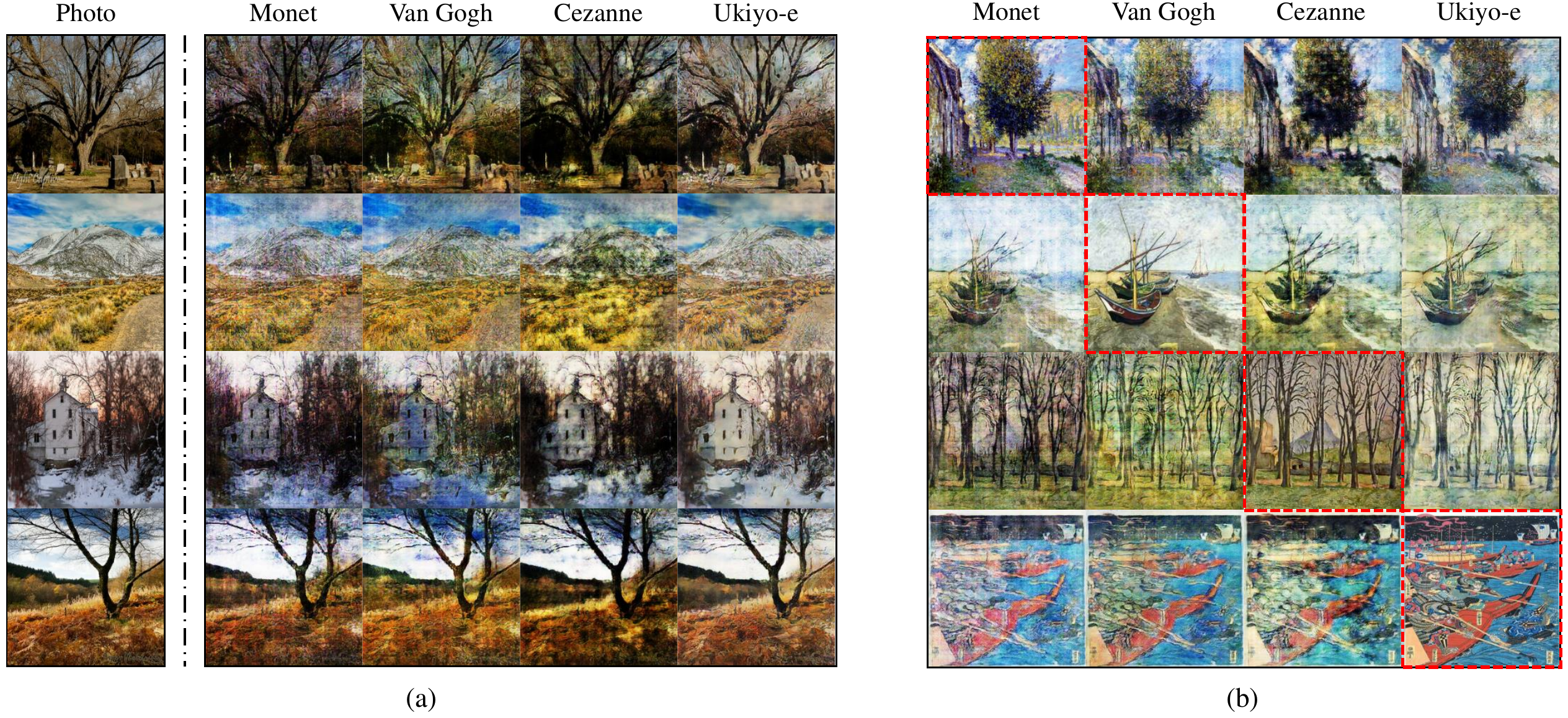}
	\vspace{-0.05cm}
	\caption{Photo to painting and painting to painting image translation results (respectively illustrated in (a) and (b)) of our method. Note that the red dotted frames in (b) denote input images.}
	\vspace{-0.4cm}
	\label{fig:8}
\end{figure*}

\subsection{Single-Attribute Transfer Evaluation}
We first evaluate our method with CycleGAN, GeneGAN, DNA-GAN, ELEGANT, StarGAN and AttGAN on the CelebA dataset for the single-attribute transfer task.
As can be seen in Fig.~\ref{fig:4}, CycleGAN is able to learn attribute information but is inefficient and generates distorted and blurred images in some cases.
GeneGAN, DNA-GAN and ELEGANT require to be specified an image of target attribute to realize attribute transfer, which is susceptible to the attribute-irrelevant information of it.
The results of GeneGAN, DNA-GAN and ELEGANT are often blurred and ghosted, indicating these three methods are incapable of decoupling image attributes and background effectively.
For StarGAN and AttGAN, both of them can capture the attribute information accurately, but some of their results are unnatural, such as ``Pale skin'' and ``Eyebrows''.
In addition, both of their results are relatively of low-resolution.
By comparison, the results of our method appear more sharp and realistic, which can fully prove that the multi-scale discriminator is conducive to generating clear images.

For quantitatively evaluate our method, we further conduct two quantitative experiments for all the methods involved.
To evaluate the attribute transfer accuracy, a classifier is designed and trained on the CelebA dataset to measure the accuracy of attribute transfer of the eight attributes by judging whether a transferred image possesses the desired attribute.
For evaluating the quality of attribute transferred images, we adopt the LPIPS metric~\cite{zhang2018unreasonable} (v0.1) to measure the perceptual similarity error between the input images and the transferred ones, which is sensitive to blurs and artifacts, and thus the lower perceptual similarity error indicates higher-resolution and more realistic transferred images.
As exhibited in Table~\ref{tab:cls} and Table~\ref{tab:iqa}, our method achieves the highest average attribute transfer accuracy and the lowest average perceptual similarity error, demonstrating the ability of our method to generate accurate and realistic results.

Our method not only fits to facial attribute transfer, but can also be extended to wide range of scenarios. We conduct an experiment of feather colors transfer on the CUB dataset and obtain promising results (See Fig.~\ref{fig:5}).

\subsection{Multiple-Attribute Transfer Evaluation}
For multiple-attribute transfer evaluation, we conduct comprehensive experiments for StarGAN, AttGAN and our method, which can learn and transfer multiple attributes simultaneously. 

As can be seen in Fig.~\ref{fig:6}, both the results of StarGAN and AttGAN are more or less distorted, such as ``Pale skin'' and ``Eyeglasses''.
Furthermore, some results are even affected by the background of images, such as ``Blond hair'', and StarGAN especially suffers from artifacts.
The reason of these undesired experimental results is that both StarGAN and AttGAN fail to decouple the information of different attributes effectively, which leads to true attributes can not be captured correctly.
By contrast, our method can generate photo-realistic and accurate results even under complex combinations of multiple attributes.
Explicitly and separately modeling image attributes and image background ensures that the attribute information and the background information are not mutually affected during generative procedure.
Different attributes are described by different means and variances of attribute latent variables, which can decorrelate those highly correlated attributes effectively.
Moreover, approaching background latent variables to uniform distributions ensures the generated images to be realistic. Conditional multi-scale discriminator is adopted to make the generated images accurate and high-resolution.
\subsection{Controllable Attribute Transfer on Manifold}
Since we adopt the idea of manifold learning for the image attribute transfer task, we can achieve a controllable and progressive transfer of attributes on the manifold. 
According to Eq.~\eqref{7}, we can control the amplitude of the attribute latent variables by controlling the value of the hyper-parameter $\theta$, and further control the intensity of attribute transfer on the manifold. The results of controllable attribute transfer are illustrated in Fig.~\ref{fig:7}, where the range of $\theta$ is $[0, 1]$. We apply six different values for $\theta$ and obtain the corresponding progressive transferring results.
\subsection{Extension to Image-to-Image Translation Task}
Our method is also a general approach for the image-to-image translation task.
By regarding the styles of images as attributes, we apply our method on the Paintings dataset and the results are illustrated in Fig.~\ref{fig:8}. Since image translation is a uniform transformation for both global and local information of an image while attribute transfer is only for local information, the results of Fig.~\ref{fig:8} contain some artifacts, but can still prove that our method is a potential architecture for a wide-range of image generation tasks.

\section{Conclusion}
We have presented a novel AME-GAN for fully-featured attribute transfer. 
The attribute latent variables and background latent variables are respectively enforced to Gaussian distributions and to uniform distributions, such that every detail in the images can be modified and adjusted by these latent variables effortlessly. 
In order to make the attribute transferred images more accurate and realistic, we developed a conditional multi-scale discriminator to distinguish generated images, which is beneficial to yielding accurate and realistic attribute transferred images. Comprehensive experimental results on three broadly used datasets demonstrate the effectiveness of the proposed approach on both attribute transfer and image translation tasks.

{\small
	\bibliographystyle{ieee}
	\bibliography{Fully-FeaturedAttributeTransfer}
}

\end{document}